\documentclass[11pt,twocolumn]{article}
\usepackage{intrepid}

\newcommand{\andromeda}{\textrm{\textsc{Andromeda}}\xspace}
\newcommand{\andromedaone}{\textrm{\textsc{Andromeda 1}}\xspace}
\newcommand{\andromedatwo}{\textrm{\textsc{Andromeda 2}}\xspace}

\graphicspath{{figures/}}

\IntrepidTitle{Evidence-Grounded Agentic Formulation \mbox{Development} in an Autonomous Laboratory}
\setIntrepidShortTitle{Agentic Formulation Development}
\IntrepidByline{Michael M. Craig, Riley J. Hickman, Yingshan Ma, R\'{e}mi Pich\'{e}-Taillefer, Christine Allen, Pauric Bannigan\textsuperscript{1}}
\IntrepidAffiliation{\textsuperscript{1}Intrepid Labs, Toronto, Canada}
\IntrepidHeroPath{fig0_first_page_takeaway.pdf}
\IntrepidAbstract{Self-emulsifying drug delivery systems (SEDDS) can improve the oral bioavailability of poorly soluble drugs, but identifying high-performing formulations remains experimentally intensive. We present \andromedatwo, an agentic system that reasons over structured in-house experimental evidence and invokes computational and experimental tools to design and execute successive formulation batches. Using a miniaturized automated laboratory at a matched budget, we benchmark it against \andromedaone, a probabilistic optimization model deployed across dozens of live development projects, and a wet-lab design-of-experiments (DoE) campaign. For paclitaxel, \andromedatwo achieved a 50\% high-performance hit rate versus 17\% for \andromedaone and 2\% for DoE, and identified 12 formulations meeting all four target product profile (TPP) objectives versus 6 and 0, respectively. Median $\mathrm{AUC}_{10\text{--}240}$ was 70.1, 12.0, and 3.5 mg$\cdot$min/mL, while maximum AUC was comparable between \andromedatwo and \andromedaone. A selected full-TPP formulation achieved an apparent effective paclitaxel loading of $19 \pm 5$\% w/w at the first FaSSIF measurement, approximately 3.3-fold higher than the 5.7\% w/w loading reported for a published paclitaxel S-SEDDS. A controlled ablation showed that access to structured in-house experimental evidence increased mean AUC by 34\%.}

\begin{document}

\IntrepidMakeTitle

\section{Introduction}
\label{sec:intro}

Self-emulsifying drug delivery systems (SEDDS) are an established strategy for
improving the oral delivery of poorly soluble drugs by enhancing apparent solubility and promoting or sustaining supersaturation following gastrointestinal dilution
\citep{pouton2006formulation,porter2007lipids}. Their development, however,
remains experimentally intensive, requiring simultaneous optimization of
oil, surfactant, cosolvent, drug loading, and, where needed, precipitation inhibitors
within strict compositional and experimental constraints. Since exhaustive exploration of such a large design space is impractical, conventional workflows typically concentrate experimental effort within selected regions through excipient screening, phase diagram construction, and design-of-experiments (DoE)-guided formulation campaigns
\citep{williams2013strategies,mu2013lipid}.

Sequential optimization methods can improve experimental efficiency by using the
results of each batch to determine what should be tested next, and autonomous
laboratories increasingly couple such algorithms to iterative
design--make--test--learn workflows
\citep{shahriari2016bo,shields2021bayesian,macleod2020selfdriving,burger2020mobile}.
However, these optimization strategies typically initialize each experimental campaign largely \textit{de novo}, learning the formulation-performance landscape primarily through experiments conducted during that campaign. For a new active pharmaceutical ingredient (API),
early proposals may therefore make limited use of the broader experimental
knowledge accumulated across previous formulation programs. This motivates a
different question: can an autonomous system reason over a laboratory's existing
experimental evidence and use that knowledge to allocate scarce wet-lab
experiments more productively?

\begin{keyresult}[Why oral paclitaxel matters]
Paclitaxel is a clinically important but exceptionally challenging candidate for oral
lipid-based formulation. Its very poor aqueous solubility requires formulations
that achieve therapeutically relevant drug loading while maintaining the drug in a solubilized or supersaturated state following
gastrointestinal dilution and minimizing precipitation. Oral exposure is further limited by biological barriers,
including intestinal P-glycoprotein efflux and first-pass metabolism, although these are outside the scope of the present
study \citep{sparreboom1997pgp}. Nevertheless, DHP107, an oral lipid-based formulation of paclitaxel,
has demonstrated noninferior efficacy relative to
intravenous paclitaxel in Phase~III clinical evaluation
\citep{kang2018dream,xu2026optimal}, establishing oral paclitaxel as a
clinically relevant formulation objective.
\end{keyresult}

We introduce \andromedatwo, an evidence-grounded agentic system for autonomous
formulation design. The system reasons over structured in-house experimental
evidence together with API, excipient, assay, and TPP
context; proposes and critiques candidate formulations; and invokes
computational and experimental tools to construct executable formulation batches. Deterministic feasibility checks enforce laboratory and compositional
constraints before robotic execution. We evaluate \andromedatwo against
\andromedaone and a physically executed DoE campaign using the same formulation
space, target product profile, assay, automated laboratory, and matched
experimental budget. \andromedaone has been used across dozens of development programs spanning diverse APIs and formulation modalities, making it an ideal point of comparison for assessing the agentic system. A reduced-evidence ablation separately tests the
contribution of accumulated in-house experimental evidence. The central question is whether evidence-grounded autonomous design can use a fixed experimental budget more effectively than conventional optimization strategies, identifying high-performing formulations more consistently.

\section{Results}
\label{sec:results}

\begin{figure*}[t]
  \centering
  \includegraphics[width=\columnwidth]{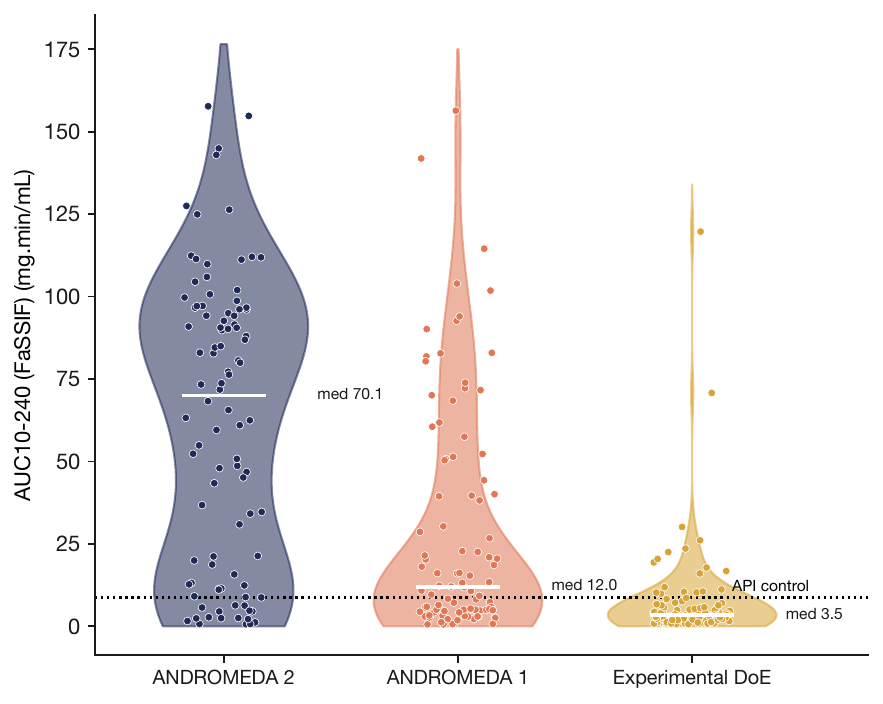}
  \caption{Paclitaxel formulation performance at a matched experimental budget. Formulation-level distributions of the area under the curve (AUC) of apparently solubilized paclitaxel
concentration in FaSSIF from 10 to 240 min achieved using the evidence-grounded agentic
platform (\andromedatwo), the optimization framework (\andromedaone) and a design of experiment (DoE) approach. Each strategy evaluated 96 unique formulations.
\andromedatwo produced an upward-shifted AUC distribution and a greater frequency of higher
performing formulations, while both \andromeda platforms generated formulations with higher
AUC values than experimental DoE.}
  \label{fig:first-page-result}
\end{figure*}

\subsection{\andromedatwo improves paclitaxel formulation performance at a matched budget}
\label{sec:results-benchmark}

\begin{figure*}[t]
  \centering
  \includegraphics[width=0.85\textwidth]{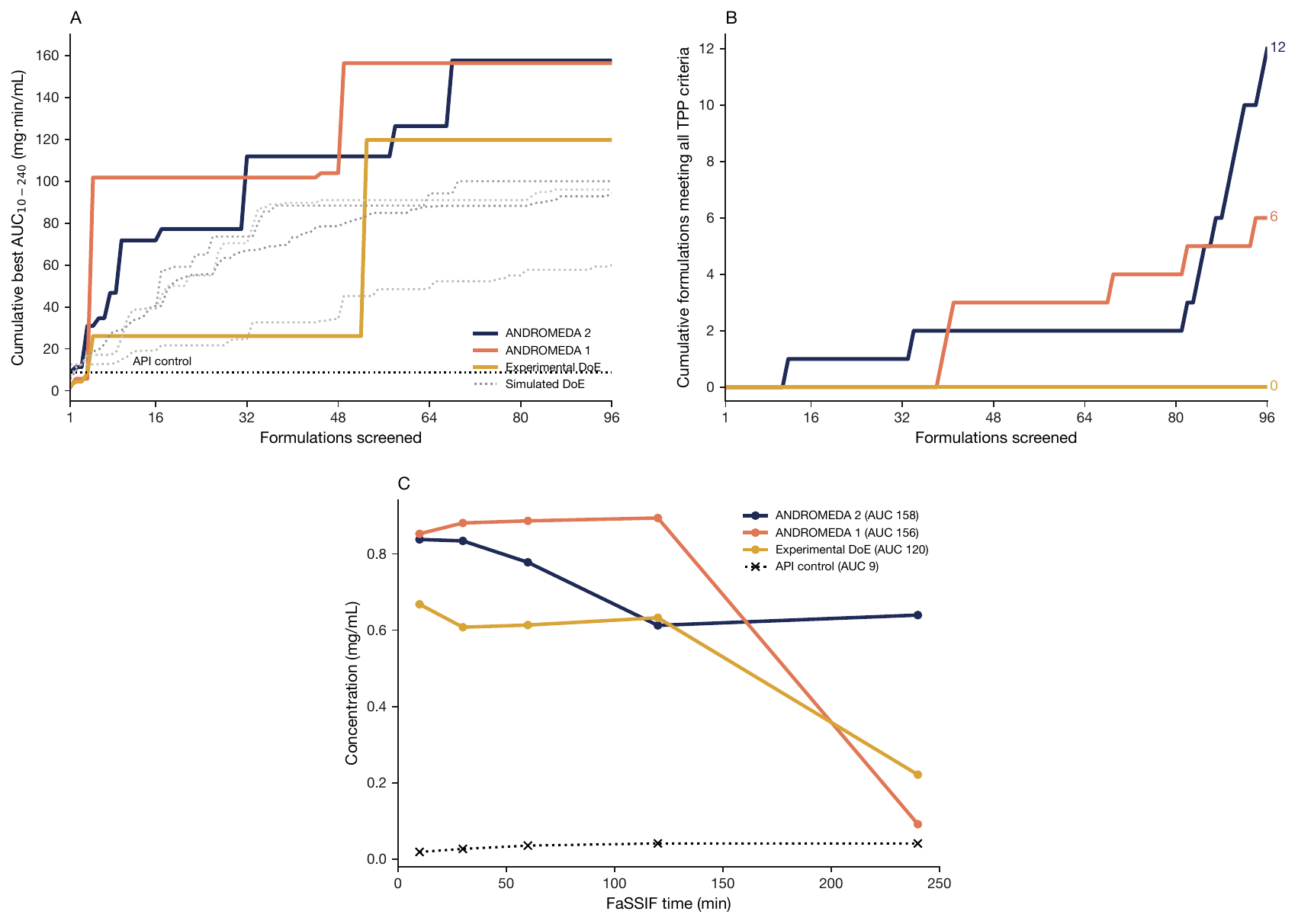}
  \caption{\textbf{Matched-budget benchmark of paclitaxel formulation performance.} All strategies
were compared at an experimental budget of 96 formulations. (A) Cumulative-best area under
the curve (AUC) of apparently solubilized paclitaxel concentration in FaSSIF from 10 to 240 min
as a function of the number of formulations evaluated. Simulated DoE trajectories (grey dotted
lines) and the unformulated-drug control (black dotted line) are included for reference. (B)
Cumulative number of formulations satisfying all four prespecified target product profile (TPP)
criteria. At 96 formulations, \andromedatwo identified 12 formulations that met all TPP objectives, compared with 6 for \andromedaone and 0 for experimental DoE. (C) In vitro
concentration–time profiles in FaSSIF for the highest AUC formulation from each experimental
strategy and the unformulated-drug control. The highest-AUC \andromedatwo formulation
maintained a high apparently solubilized paclitaxel concentration through 240 min, whereas the
highest AUC \andromedaone formulation reached a comparable peak concentration but declined
after 120 min, consistent with precipitation.}
  \label{fig:benchmark}
\end{figure*}

We compared \andromedatwo with \andromedaone and a physically executed DoE campaign using the same miniaturized laboratory and a matched budget of 96 unique formulations per strategy. \andromedatwo, \andromedaone, and experimental DoE attained median $\mathrm{AUC}_{10\text{--}240}$ values of 70.1, 12.0, and 3.5~mg$\cdot$min/mL,
respectively, and high-AUC hit rates of 50\%, 17\%, and 2\%
(\figref{fig:first-page-result}; Table~\ref{tab:headline-summary}; \figref{fig:benchmark}A). The maximum
observed AUC was similar for \andromedatwo and \andromedaone (157.7 vs
156.4~mg$\cdot$min/mL).
A descriptive formulation-level Mann--Whitney comparison is reported in the
Supplementary Material (Section~\ref{sec:si-mannwhitney}).

\andromedatwo produced 12 formulations that met all four TPP criteria, compared
with 6 for \andromedaone and 0 for experimental DoE
(\figref{fig:benchmark}B; \figref{fig:tppdef}). The highest-AUC formulation
met only two of the four TPP objectives, so maximum AUC is not equivalent to a balanced
formulation. \andromedatwo's best formulation maintained apparent solubilized
paclitaxel through 240~min and reached an AUC $\sim$18$\times$ the unformulated control drug (\figref{fig:benchmark}C).

\begin{table*}[t]
\centering
\caption{Summary of paclitaxel formulation-selection performance at a
matched 96-formulation experimental budget.}
\label{tab:headline-summary}

\vspace{0.5em}

\resizebox{\textwidth}{!}{%
\begin{tabular}{lccccc}
\toprule
\textbf{Arm}
& \textbf{$N$}
& \shortstack{\textbf{Best}\\\textbf{AUC}}
& \shortstack{\textbf{Median}\\\textbf{AUC}}
& \shortstack{\textbf{High-AUC hits}\\\textbf{$n$ (\%)}}
& \shortstack{\textbf{Formulations to}\\\textbf{first full TPP}}
\\
\midrule

\andromedatwo
& 96
& \textbf{157.7}
& \textbf{70.1}
& \textbf{48 (50\%)}
& \textbf{12}
\\

\andromedaone
& 96
& 156.4
& 12.0
& 16 (17\%)
& 39
\\

Experimental DoE
& 96
& 119.7
& 3.5
& 2 (2\%)
& No pass
\\

Reduced-evidence \andromedatwo
& 96
& 123.3
& 46.1
& 30 (31\%)
& 53
\\

\bottomrule
\end{tabular}%
}

\vspace{1.5mm}

\begin{minipage}{0.96\linewidth}
\scriptsize
AUC denotes $\mathrm{AUC}_{10\text{--}240}$ in FaSSIF.
High-AUC hits are formulations reaching the pooled upper-quartile AUC threshold.
``Formulations to first full TPP'' denotes the number of formulations screened
before the first formulation satisfying all four TPP criteria was identified.
Experimental DoE identified no full-TPP formulation within the
96-formulation campaign.
\end{minipage}

\end{table*}

\subsection{Structured in-house evidence improves \andromedatwo performance}
\label{sec:results-ablation}

\begin{figure*}[t]
  \centering
  \includegraphics[width=0.5\textwidth]{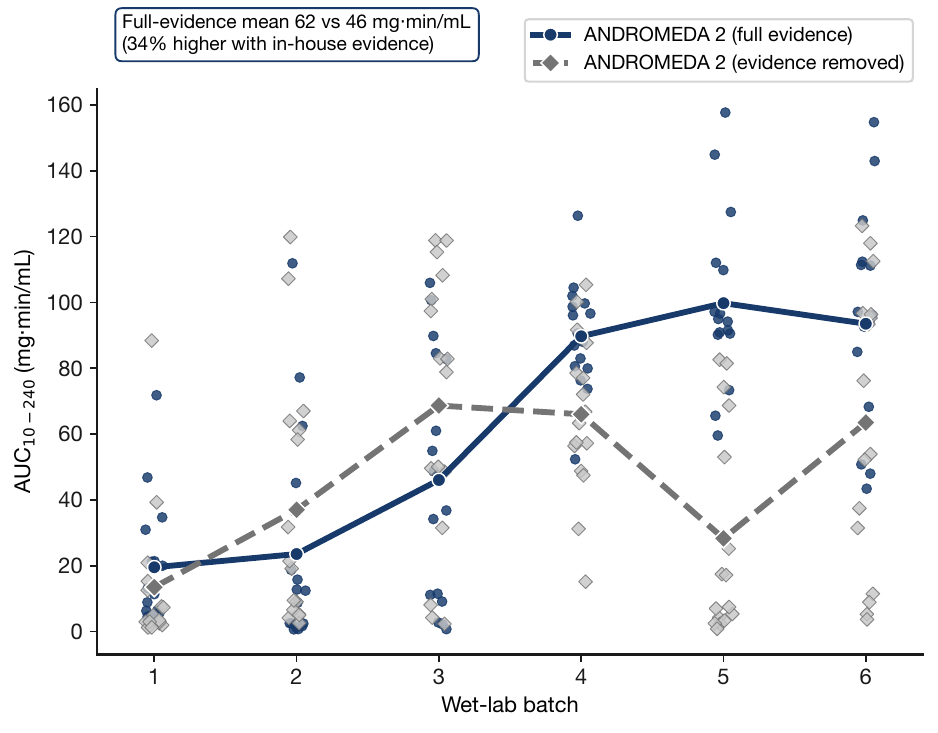}
  \caption{\textbf{Effect of in-house evidence on \andromedatwo performance across paclitaxel batches.}
    Per-formulation area under the curve of apparently solubilized paclitaxel concentration in FaSSIF
from 10 to 240 min is shown across six sequential wet-lab batches for \andromedatwo with access
to its full structured in-house evidence and for the reduced-evidence ablation in which this
evidence was withheld. Both configurations received the experimental results generated during
their respective campaigns. Points represent individual formulations, and lines connect batch
means. The full-evidence configuration achieved a higher campaign-wide mean AUC than the
reduced-evidence configuration (62.0 versus 46.2 mg·min/mL; 34\% higher), showed more
consistent improvement across batches, and maintained its gains in later batches. The reduced-
evidence configuration matched or exceeded the full-evidence configuration in some early
batches but did not maintain this performance.}
  \label{fig:ablation}
\end{figure*}

To isolate the contribution of Intrepid's structured in-house experimental evidence, we compared full-evidence \andromedatwo with an otherwise matched reduced-evidence configuration on paclitaxel (\figref{fig:ablation}). Access to historical in-house experimental evidence was withheld in the reduced-evidence condition, while the formulation objective, executable design space, TPP, feasibility constraints, batch size, and wet-lab workflow were held constant. Both configurations continued to receive the paclitaxel measurements generated during their respective campaigns after each batch. 

Across the full campaign, mean $\mathrm{AUC}_{10\text{--}240}$ was 62.0 mg$\cdot$min/mL with full evidence versus 46.2 mg$\cdot$min/mL with the evidence withheld, corresponding to a 34\% higher mean AUC. The high-AUC hit rate was similarly higher with full evidence at 50\% versus 31\%. The batch-resolved trajectories suggest that this difference was not simply due to a stronger initial batch. 

The reduced-evidence configuration matched or out-performed full-evidence \andromedatwo in early batches but subsequently regressed, whereas full-evidence \andromedatwo improved across successive batches and sustained its performance. This trajectory is consistent with historical evidence supporting more reliable interpretation and use of newly generated experimental results, rather than simply providing an initial advantage. The reduced-evidence system nevertheless continued to generate valid, experimentally executable SEDDS formulations, indicating that the ablation primarily affected candidate selection rather than formulation feasibility.

\subsection{Benchmarking \andromedatwo against published oral paclitaxel formulations}
\label{sec:results-literature}
Published oral paclitaxel formulations provide useful context for the
formulations identified by \andromedatwo, although differences in formulation
type, experimental endpoint, and study setting preclude direct head-to-head
comparison. In particular, the FaSSIF $\mathrm{AUC}_{10\text{--}240}$ measured
here should not be interpreted as oral bioavailability or systemic exposure.
A compact literature comparison is reported in the Supplementary Material
(Table~\ref{tab:ptx_literature_benchmark}).

For the selected formulation meeting all four TPP objectives, the apparent
effective paclitaxel loading estimated from the first FaSSIF measurement
(10~min) was $19 \pm 5$\% w/w across three replicates. This is approximately
3.3-fold higher than the 5.7\% w/w paclitaxel loading reported for the
S-SEDDS formulation of Gao et al.~\cite{gao2003ssedds}, and more than
an order of magnitude higher than the loadings reported for selected clinically
evaluated lipid-based oral paclitaxel formulations. However, these values provide
formulation context rather than a direct quantitative ranking because loading
and measurement methods differ across studies.

\section{Discussion}
\label{sec:discussion}

This study shows that evidence-grounded agentic design can substantially improve the experimental efficiency of SEDDS formulation development. At a matched budget of 96 paclitaxel formulations per strategy, \andromedatwo achieved a 50\% high-AUC hit rate, compared with 17\% for \andromedaone and 2\% for the physically executed DoE campaign, and identified 12 formulations meeting all four TPP criteria, compared with 6 and 0, respectively. Notably, \andromedatwo and \andromedaone reached comparable maximum AUC values, indicating that the agentic system's main advantage was not identifying a higher isolated optimum, but producing substantially more high-performing formulations across the experimental campaign. These findings extend a growing literature on self-driving laboratories and data-driven optimization in chemistry and materials science \citep{burger2020mobile,macleod2020selfdriving,shields2021bayesian,shahriari2016bo}, as well as machine-learning approaches to drug formulation design \citep{bannigan2021ml,bannigan2023lai,reker2021computationally}. Here, the distinguishing contribution is the coupling of autonomous experimentation with evidence-grounded agentic decision-making to identify high-performing formulations more frequently within a fixed wet-lab budget than either probabilistic optimization or DoE.

The evidence-ablation and composition analyses provide complementary insight into how the \andromedatwo advantage emerged. Access to structured in-house experimental evidence increased mean paclitaxel AUC by 34\% and the high-AUC hit rate from 31\% to 50\%. In addition, \andromedatwo achieved a 49\% hit rate within the high-performing Type~IIIB/IV region, compared with 29\% for \andromedaone, indicating that its advantage was not explained solely by identifying a favourable region of composition space. More importantly, the reduced-evidence configuration remained competitive in early batches but subsequently regressed, whereas full-evidence \andromedatwo improved across successive batches and sustained its performance. Both configurations received the experimental results generated during their respective campaigns; the divergent trajectories are therefore consistent with historical evidence supporting more effective interpretation and use of newly generated measurements, rather than simply strengthening the initial proposal. The ablation does not by itself attribute the full \andromedatwo--\andromedaone performance difference to historical evidence, but demonstrates that access to this evidence materially contributes to \andromedatwo performance. The reduced-evidence system continued to generate valid, experimentally executable formulations, further suggesting that the principal effect of the ablation was on candidate selection rather than formulation feasibility. This finding is consistent with the evidence-ablation effect we previously observed for clofazimine \citep{craig2026agentic}.
From a formulation science perspective, \andromedatwo concentrated experimental effort within high-performing regions of the formulation design space while identifying multiple lead candidate formulations rather than a single isolated composition. Vitamin~E~TPGS was prominent among these high-performing compositions, consistent with its established paclitaxel-solubilizing properties. Vitamin~E~TPGS also exhibits independent effects on P-glycoprotein-mediated paclitaxel transport, which may prove advantageous in future \textit{in vivo} studies. Since intestinal transport was not measured in the present study, these results support only the physicochemical performance of the formulations in FaSSIF; nevertheless, identifying formulation chemistry relevant to both paclitaxel solubilization and a known biological barrier to oral delivery is encouraging. The apparent effective paclitaxel loading of formulations meeting all four TPP criteria also provides a useful point of comparison with previously reported oral lipid-based paclitaxel systems (Table~\ref{tab:ptx_literature_benchmark}), although differences in experimental endpoints preclude direct ranking.

Several limitations define the scope of these findings. The FaSSIF assay measures the ability of a formulation to maintain paclitaxel in an apparently solubilized state and does not capture digestion, intestinal permeability or metabolism, or \textit{in vivo} performance; digestion-coupled testing and ultimately pharmacokinetic studies will therefore be required to determine whether the observed formulation's advantages translate to oral exposure. In addition, each strategy was evaluated in a single independently initialized campaign, limiting campaign-level statistical inference and motivating replicate autonomous campaigns to establish reproducibility. The experimental DoE comparator also explored the shared composition space directly, without the expert pre-screening and prioritization that may accompany conventional formulation development, and should be interpreted in that context. Finally, prospective evaluation across chemically distinct APIs is needed to establish the generality of the evidence-grounded advantage. Together, these studies would test whether the improved allocation of experimental effort observed here persists across campaigns, APIs, and increasingly translational endpoints.

More broadly, these findings point toward a model of formulation development in which experimental data are not consumed within individual programs but accumulated as reusable evidence that improves subsequent autonomous decision-making. If this advantage generalizes across APIs and formulation modalities, evidence-grounded autonomous laboratories could make formulation development progressively more efficient as institutional experimental knowledge grows.

\section{Methods}
\label{sec:methods}

\subsection{Autonomous SEDDS laboratory}
\label{sec:lab}

All campaigns used the same miniaturized design--build--test platform. Each
wet-lab batch comprised 16 SEDDS preconcentrates prepared by liquid-handling
robots and evaluated by automated dispersion/dissolution, dynamic light
scattering (droplet size and polydispersity index [PDI]), HPLC drug quantification,
and structured data capture. All experimental arms used the same operators,
instruments, assay timing, data-processing workflow, and dissolution protocol.

Formulations were evaluated in a two-stage biorelevant dissolution assay: an
initial gastric-fluid stage followed by transfer into fasted-state simulated
intestinal fluid (FaSSIF) \citep{galia1998evaluation}. Apparent solubilized-drug
concentrations were quantified by HPLC analysis of sampled aliquots at 10, 30, 60, 120
and 240 min after the intestinal-fluid transfer. The reported endpoint was the
apparent concentration of solubilized drug in the FaSSIF stage. This endpoint
captures the ability of a formulation to disperse and maintain drug in an
apparently solubilized state under biorelevant conditions. It was used as the
principal high-throughput measure of formulation performance and not
interpreted as a surrogate for absorbed dose or oral bioavailability.

Drug loading was specified as a formulation-design input rather than measured
in the undiluted preconcentrate. An apparent effective loading was therefore
estimated retrospectively by multiplying the nominal loading by the maximum
fraction of the nominal API dose observed in the dispersed phase by HPLC analysis. This
operational estimate is used only for comparison with literature formulations
and is not a direct measurement of preconcentrate drug content.

\subsection{\andromedaone: probabilistic optimization}
\label{sec:bo}

\andromedaone is a sequential probabilistic-optimization platform and the
non-agentic comparator in this study. Candidate formulations are drawn from a
predefined, constrained composition space and evaluated against the same TPP used by \andromedatwo. At each iteration, probabilistic
models relate composition to measured outcomes and guide selection of the next
16-formulation batch, balancing promising regions of design space against
formulations whose performance remains uncertain. Newly measured dissolution,
droplet size, PDI and HPLC data are incorporated before the next cycle. The
matched paclitaxel campaign comprised six batches (96 formulations).

\subsection{\andromedatwo: agentic formulation design}
\label{sec:architecture-method}

\begin{figure*}[t]
  \centering
  \includegraphics[width=\textwidth]{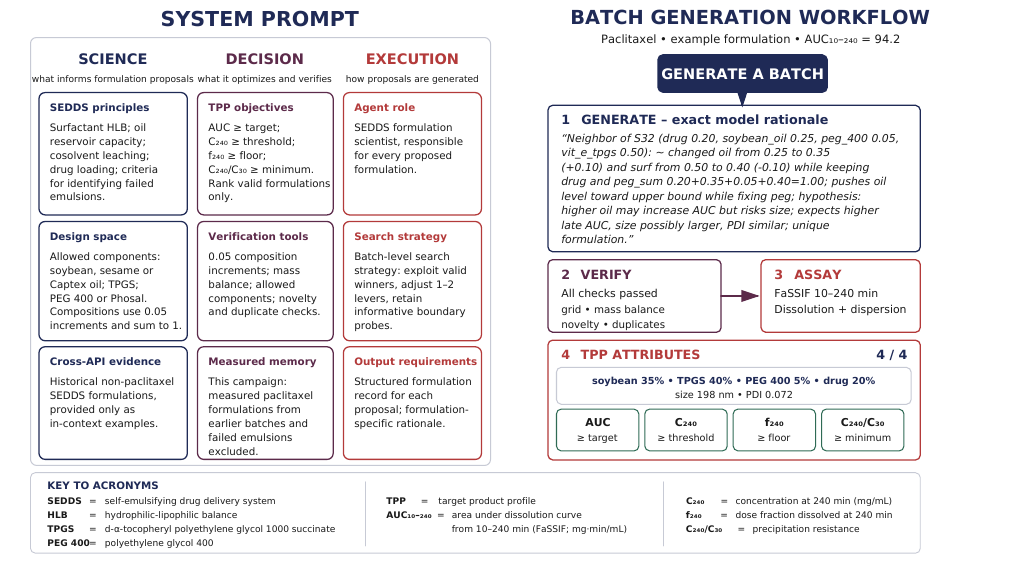}
  \caption{\textbf{System prompt structure and representative batch-generation workflow for Andromeda 2.} The system prompt organizes formulation knowledge, optimization objectives, and execution requirements into science, decision, and execution components. During batch generation, \andromedatwo uses this context to propose candidate formulations with model-generated rationales, verifies compositional and feasibility constraints, and submits valid formulations for experimental evaluation. Measured formulation performance is then evaluated against the predefined target product profile (TPP) attributes, including dissolution AUC, late-time concentration, fraction dissolved, and precipitation resistance. The example shown illustrates a representative formulation from a generated batch rather than the complete batch-level output.}
  \label{fig:architecture}
\end{figure*}

\andromedatwo is an evidence-grounded agentic system that orchestrates models
and other computational capabilities as tools (\figref{fig:architecture}).
Proposal, critique, review, planning, and constraint-checking operate over a
shared formulation context. The system can invoke the probabilistic models used
by \andromedaone, together with additional models, to construct 16-formulation
batches with scientific rationale and platform-compatible composition records.

A deterministic feasibility engine then enforces the hard constraints required
for wet-lab execution, including valid excipient selections, compatibility with
the permitted composition grid, sum-to-one compositional constraints,
uniqueness relative to previously tested formulations, intra-batch
distinctiveness, and robotic readiness. This separation distinguishes scientific
search from experimental implementation: the agentic layer directs exploration,
while the engineering layer ensures that submitted compositions are executable.

A formulation scientist reviewed each proposed batch before execution as a
safety and feasibility check that the batch was scientifically reasonable and
experimentally executable. This was not a redesign step: the reviewer did not
author, substitute, or re-tune the proposed compositions.

With no measurements yet available from the current campaign, the first batch explored several first-principles hypotheses about which excipient combinations could satisfy the TPP. Subsequent batches received all preceding
measurements, and the system preferentially explored perturbations of
high-performing formulations while retaining limited broader exploration. No
model retraining occurred between batches; sequential adaptation was achieved
by updating the agent's working context with newly generated experimental
evidence.

Both platforms are fully in-house. Language models, probabilistic models and
supporting models are developed, trained and operated on Intrepid
infrastructure, using Intrepid experimental data exclusively. Neither platform
depends on external commercial model application programming interfaces.

\subsection{Evidence grounding and ablation}
\label{sec:evidence-layers}

A key distinction between the two Andromeda versions is the evidence available
at the initiation of an experimental campaign. \andromedaone does not have access to
historical formulation data. Instead, each campaign begins with an initial
design batch, after which the planner learns the formulation--performance
landscape from measurements generated during that campaign and uses these
observations to guide subsequent batches within a probabilistic-optimization
framework. In contrast, \andromedatwo is grounded from the outset in Intrepid's
structured in-house formulation evidence, comprising prior compositions,
measured dissolution, dispersion, size and PDI outcomes, excipient behaviour,
and platform-specific measurement history. Importantly, this historical
evidence contained no paclitaxel formulations, such that \andromedatwo could
draw on broader formulation knowledge without access to prior experimental
results for the specific API evaluated here. It can reason over this accumulated evidence
during proposal, critique, planning, and tool use, while also incorporating
observations generated during the ongoing campaign.

To evaluate transfer to previously unseen APIs, all historical formulation data
for the target API were excluded at the initiation of each campaign. The systems
were not trained on in-house experimental formulation data for paclitaxel. After
each wet-lab batch, newly generated target-API measurements were returned to
the system to guide subsequent batch selection. The campaigns therefore assess
whether knowledge accumulated across Intrepid's broader laboratory experience
can support formulation design for a new API, followed by sequential adaptation
to measurements generated during the campaign.

To isolate the contribution of this accumulated evidence, paclitaxel was also
run in a reduced-evidence \andromedatwo configuration that withheld historical
in-house experimental evidence while holding constant the formulation
objective, TPP, executable design space, feasibility constraints, batch size
and wet-lab workflow. Both configurations received the paclitaxel measurements
generated during their respective campaigns after each batch.

\subsection{Comparators and paclitaxel design task}
\label{sec:doe}

Paclitaxel was selected as a hard-API stress test (high molecular weight and
very low aqueous solubility) and evaluated with \andromedatwo, \andromedaone,
reduced-evidence \andromedatwo, and a physically executed experimental DoE approach. Each
arm received a matched budget of six 16-formulation batches (96 unique
formulations).

The experimental DoE was designed by a formulation scientist using commercial
statistical software supporting custom and response-surface designs over the
same composition parameters and hard constraints, and executed end-to-end on
the same platform and FaSSIF assay. It explored the shared composition grid
directly rather than beginning from a separate excipient solubility
pre-screen. 

Simulated non-adaptive and sequential adaptive DoE strategies were additionally
evaluated against a validated \textit{in-silico} emulator of the paclitaxel formulation
landscape. Emulator validation and those results are reported in the
Supplementary Material (Section~\ref{sec:emulator},
Table~\ref{tab:oracle-validation}).

\subsection{Metrics and statistical analysis}
\label{sec:scoring}

The primary endpoint is trapezoidal $\mathrm{AUC}_{10\text{--}240}$ in FaSSIF.
Cumulative-best AUC versus formulations screened describes how rapidly
high-performing formulations appear. Mean and median AUC describe the
distribution of performance, which is distinct from peak AUC: a method may
identify a single exceptional formulation while producing substantially lower
typical candidates. A high-AUC hit is a formulation whose
$\mathrm{AUC}_{10\text{--}240}$ reaches the pooled upper quartile of all
measured paclitaxel formulations. Because that threshold is relative,
comparisons are also reported against the pool-independent TPP.

\begin{figure*}[t]
  \centering
  \includegraphics[width=\textwidth]{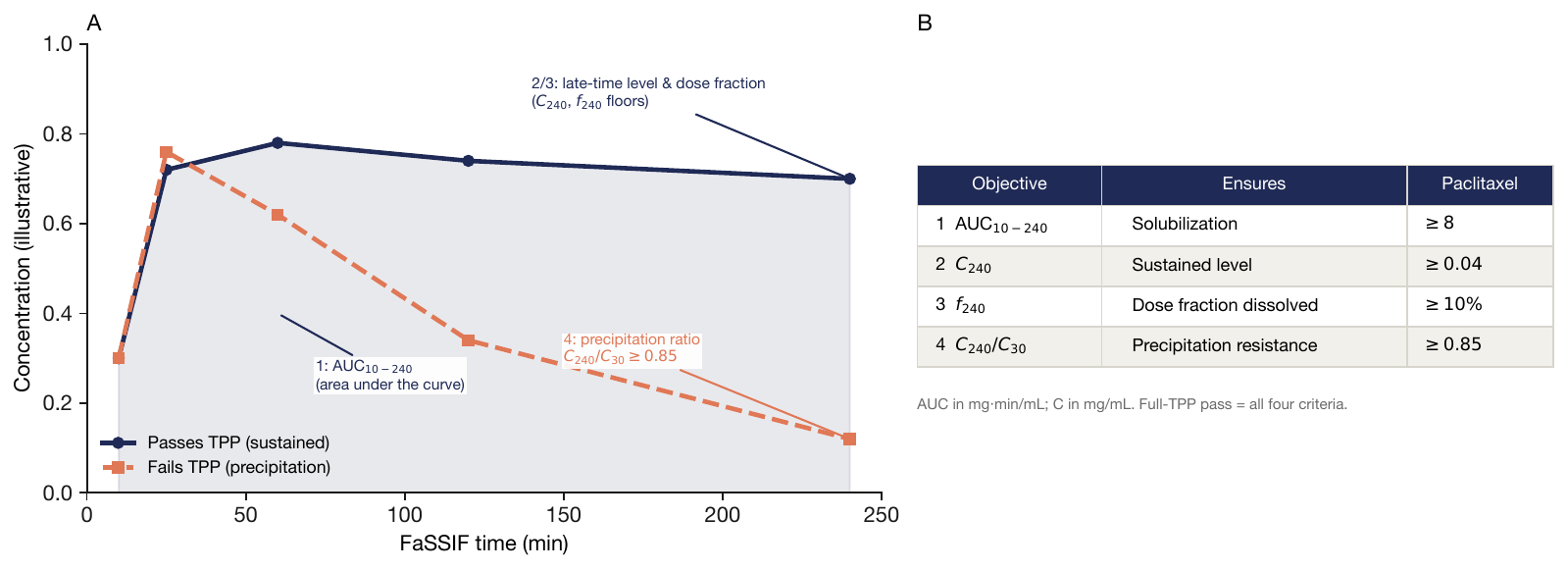}
  \caption{\textbf{Target product profile (TPP) definition.} (\textbf{A}) An
    illustrative formulation that meets the TPP (sustained) and one that does
    not (precipitating), annotated with the four TPP objectives.
    (\textbf{B}) Acceptance objectives for paclitaxel. A formulation meets the
    full TPP only if all four objectives are satisfied.}
  \label{fig:tppdef}
\end{figure*}

Formulation usability is summarized with a four-attribute TPP
(\figref{fig:tppdef}): a minimum dissolution AUC, a late-time concentration
floor $C_{240}$, a late-time fraction-dissolved floor $f_{240}$ (which
normalizes for drug loading), and a precipitation-resistance ratio
$C_{240}/C_{30} \geq 0.85$. A formulation meets the full TPP only if all four
objectives are satisfied. Thresholds are API-specific
(\figref{fig:tppdef}B). Peak AUC, the number of objectives met, and full-TPP
passage are not interchangeable.

Pairwise distributional comparisons use two-sided Mann--Whitney U tests. Hit rates are reported with 95\% confidence intervals calculated using percentile
bootstrapping. We lead with effect sizes (medians, hit
rates, and counts of formulations meeting all TPP objectives). All
formulation-level tests and bootstrap intervals are treated as descriptive
because observations within adaptive campaigns are not statistically
independent; no campaign-level inferential testing was performed.

\clearpage

\bibliographystyle{unsrtnat}
\bibliography{references}

\clearpage

\onecolumn
\raggedbottom
\appendix
\setlength{\textfloatsep}{8pt plus 2pt minus 4pt}
\setlength{\intextsep}{8pt plus 2pt minus 4pt}
\setlength{\floatsep}{8pt plus 2pt minus 4pt}
\captionsetup{skip=4pt, hypcap=false}

\section{Supplementary Material}
\label{sec:supplementary}

\subsection{Descriptive Mann--Whitney comparison of \andromedatwo and \andromedaone}
\label{sec:si-mannwhitney}

A formulation-level Mann--Whitney comparison between \andromedatwo and
\andromedaone yielded $p=9.6\times10^{-8}$. This value is treated as descriptive
rather than inferential because observations within each adaptive campaign are
not independent: selection of later formulations depends on the measured
performance of formulations tested in earlier batches. The nominal sample size
of 96 formulations per strategy therefore overstates the effective sample size.
Interpretation of the matched-budget result in the main text therefore rests
primarily on the magnitude of the distributional shift and the observed hit
rates.

\subsection{Formulation composition and the LFCS taxonomy}
\label{sec:si-lfcs}

\begin{figure}[!htbp]
  \centering
  \includegraphics[width=\textwidth]{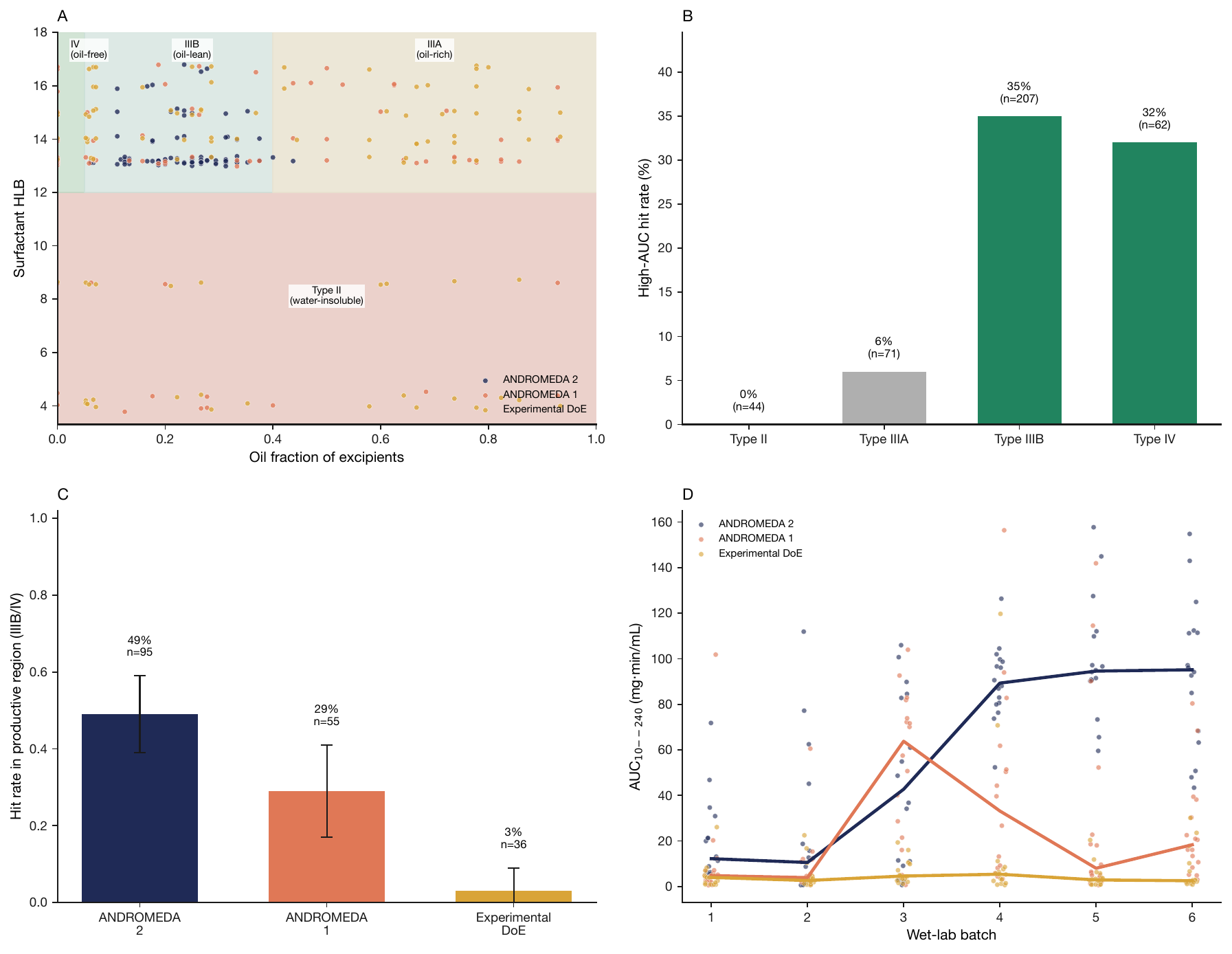}
  \caption{\textbf{Formulation composition using the LFCS taxonomy (paclitaxel).}
    (\textbf{A}) Composition space (oil fraction versus surfactant HLB) with
    LFCS Type regions shaded and formulations coloured by method. (\textbf{B})
    High-AUC hit rate by LFCS Type; the water-soluble-surfactant Type~IIIB/IV
    are productive. (\textbf{C}) Hit rate within the productive region by method,
    with bootstrap 95\% confidence intervals. (\textbf{D}) Batch progression:
    each point is a formulation (batch on the $x$-axis, AUC on the $y$-axis),
    with per-method batch medians.}
  \label{fig:composition}
\end{figure}

Formulation development decisions depend not only on the final formulations identified but also on how the method employed allocates its experimental budget across composition space. To characterize these exploration patterns, we classified each formulation using the Lipid Formulation Classification System (LFCS), a widely used compositional framework for lipid-based formulations
\citep{pouton2006formulation}.  The LFCS distinguishes formulations according to their
relative proportions of oil, water-insoluble surfactant (typically HLB less
than 12), water-soluble surfactant (typically HLB greater than 12), and
hydrophilic cosolvent (\figref{fig:composition}A).

Within the design space, formulations were classified as Type II, comprising
oil and water-insoluble surfactant without water-soluble components; Type IIIA,
containing substantial oil together with water-soluble surfactant and optionally
cosolvent; Type IIIB, containing less oil and a greater proportion of
water-soluble components; or Type IV, an oil-free system based predominantly on
water-soluble surfactants and cosolvents (\figref{fig:composition}A). No Type I
oil-only formulations were included in the evaluated design space. For each method, we report the distribution of formulations across LFCS types, the hit rate within the productive region, and the evolution of hit rate across experimental batches.

Across the tested paclitaxel formulations, the high-performance hit rate varied
markedly by LFCS type: 0\% for Type~II, 6\% for Type~IIIA, 35\% for Type~IIIB,
and 32\% for Type~IV (\figref{fig:composition}B). Thus, within the evaluated
design space, high-performing formulations were concentrated in the
Type~IIIB/IV region, characterized by water-soluble surfactants and low or no
oil content. \andromedatwo directed most of its experimental budget toward this
region and largely avoided the unproductive Type~II systems, whereas the
experimental DoE allocated a substantial fraction of its budget to Type~II
formulations.

The advantage of \andromedatwo was not attributable solely to identifying the
productive LFCS region. Among Type IIIB/IV formulations, \andromedatwo achieved a
49\% hit rate (bootstrap 95\% CI $[0.39, 0.59]$), compared with 29\% for
\andromedaone and 3\% for experimental DoE (\figref{fig:composition}C).
\andromedatwo therefore both concentrated its search within the most productive
region and selected higher-performing candidates within that region.

The composition trajectory also demonstrated adaptation over the course of the
campaign (\figref{fig:composition}D). \andromedatwo began with broad chemical
exploration sampling 12 distinct oil-surfactant families in its first batch. In
subsequent batches it increasingly concentrated on productive Type IIIB/IV
formulations as experimental measurements accumulated and per-formulation AUC
increased. DoE showed little corresponding improvement across batches. Although
\andromedatwo sampled fewer distinct oil-surfactant families overall than
experimental DoE (16 versus 90), it generated substantially more
high-performing formulations. This focused search did not collapse onto a
single composition family: \andromedatwo retained several productive families,
providing chemically distinct alternatives for subsequent formulation
development.

\subsection{Literature comparison of oral paclitaxel formulations}
\label{sec:si-ptx-literature}

The table below is offered as context rather than as a ranking of formulations.
Endpoints differ across studies, and the FaSSIF AUC used in the main text
should not be equated with oral bioavailability or systemic exposure.

\vspace{0.5\baselineskip}
\noindent
\begin{minipage}{\textwidth}
\centering
\small
\setlength{\abovecaptionskip}{2pt}
\captionof{table}{Selected published lipid-based oral paclitaxel formulations compared
  with \andromedatwo. Endpoints differ across studies and are shown for context
  rather than as direct head-to-head comparisons.}
\label{tab:ptx_literature_benchmark}
\begin{tabular}{p{3.6cm}|p{3.6cm}|p{7.0cm}}
\toprule
\textbf{Formulation} &
\textbf{PTX loading (\% w/w)} &
\textbf{Formulation / translational result} \\
\midrule

\andromedatwo &
Up to 25 (nominal)$^{a}$ &
12/96 full-TPP passes; high apparent solubilized PTX maintained in FaSSIF
through 240~min. \\

\citet{gao2003ssedds}, S-SEDDS &
5.7 &
Precipitation-inhibiting polymer improved maintenance of PTX after dilution;
subsequent rat PK demonstrated oral exposure. \\

\citet{veltkamp2006smeof}, SMEOF\#3 &
1.6$^{c}$ &
Clinical oral PTX formulation; evaluated in cancer patients with
cyclosporine~A. \\

DHP107 \citep{hong2007dhp107,xu2026optimal} &
$\sim$1$^{c}$ &
Clinically validated lipid-based oral PTX formulation$^{b}$; Phase~III
development without co-administered P-gp inhibitor. \\
\bottomrule
\end{tabular}

\vspace{0.4em}
\begin{minipage}{\linewidth}
\footnotesize
$^{a}$Nominal drug loading is a formulation-design input in the present study
and should not be interpreted as directly measured drug content or as effective
solubilized loading. 

$^{b}$DHP107 is included as a translational lipid-based oral paclitaxel
benchmark and should not be described as a conventional SEDDS.

$^{c}$Published loadings that were not originally reported as \% w/w are
converted here for comparison. Gao et al.\ reported 57~mg/g paclitaxel,
which is 5.7\% w/w. Veltkamp et al.\ reported 1.6\% w/v (160~mg in 10~mL);
their composition table sums to 100~g per 100~mL, so 1.6\% w/v is numerically
equivalent to 1.6\% w/w under that accounting. DHP107 is reported as
10~mg/mL (approximately 1\% w/v); the \% w/w value assumes a formulation
density of $\sim$1~g/mL and should be treated as approximate.
\end{minipage}
\end{minipage}

\subsection{\textit{In silico} emulator validation}
\label{sec:emulator}
The simulated DoE strategies shown in \figref{fig:benchmark}A were evaluated
using an \textit{in-silico} emulator of the paclitaxel formulation landscape
trained on in-house data. On held-out formulations the emulator recovered the
measured performance ordering (Spearman 0.89) and enriched top-quartile
formulations 3.02-fold relative to random selection, against a ceiling of 4.0
for perfect selection (Table~\ref{tab:oracle-validation}). Under these
conditions, neither non-adaptive nor sequential adaptive DoE reached the
upper-tail AUC values measured for the \andromeda strategies. This is
consistent with the experimental DoE campaign, which likewise identified no
formulations in that range.

These simulated arms are reported as directional context rather than as a
matched comparator. The emulator was validated on random held-out splits; its ranking accuracy is lower for oil--surfactant
composition families not represented in that data, and the simulated result
should be weighted accordingly.

\vspace{0.5\baselineskip}
\noindent
\begin{minipage}{\textwidth}
\centering
\setlength{\abovecaptionskip}{2pt}
\captionof{table}{\textit{In silico} emulator validation for paclitaxel:
    held-out ranking skill, error, and top-quartile enrichment. MAE in mg·min/mL. Top-quartile enrichment is relative to a chance
baseline of 1.0 (ceiling 4.0 for perfect selection).
    }
\label{tab:oracle-validation}
\begin{tabular}{lrrrr}
\toprule
API & Test Spearman & Test MAE & Top-Q enrich. \\
\midrule
Paclitaxel & 0.89 & 12.5 & 3.02 \\
\bottomrule
\end{tabular}
\end{minipage}

\par
\vspace{0.75\baselineskip}












\end{document}